\documentclass[wcp]{jmlr}

\usepackage{longtable}% for long tables

\usepackage{booktabs}
\usepackage[load-configurations=version-1]{siunitx} % newer version
\usepackage{siunitx}
\usepackage{natbib}

\usepackage{picinpar}
\usepackage{graphicx}
\usepackage{comment}
\usepackage{amsmath,amssymb}
\usepackage{bbm}
\usepackage{color}
\usepackage{url}
\usepackage{multirow}
\usepackage{booktabs}
\def\etal{\emph{et al}.}

\makeatletter
\let\Ginclude@graphics\@org@Ginclude@graphics 
\makeatother

\jmlrvolume{}
\jmlryear{2021}
\jmlrworkshop{ACML 2021}

\title[Domain Adaptive YOLO for One-Stage Cross-Domain Detection]{Domain Adaptive YOLO for One-Stage\\ Cross-Domain Detection}

  \author{\Name{Shizhao Zhang} \Email{zhang\_shizhao@sjtu.edu.cn}\\
  \Name{Hongya Tuo} \Email{huohy@sjtu.edu.cn}\\
  \Name{Zhongliang Jing} \Email{zljing@sjtu.edu.cn}\\
  \addr {Shanghai Jiao Tong University, Shanghai, China}\\
  \Name{Jian Hu} \Email{Jianhu18@sjtu.edu.cn}\\
  \addr {Baidu Inc., Shanghai, China}}

\begin{document}

\maketitle

\begin{abstract}
Domain shift is a major challenge for object detectors to generalize well to real world applications. Emerging techniques of domain adaptation for two-stage detectors help to tackle this problem. However, two-stage detectors are not the first choice for industrial applications due to its long time consumption. In this paper, a novel Domain Adaptive YOLO (DA-YOLO) is proposed to improve cross-domain performance for one-stage detectors. Image level features alignment is used to strictly match for  local features like texture, and loosely match for global features like illumination. Multi-scale instance level features alignment is presented to reduce instance domain shift effectively , such as variations in object appearance and viewpoint. A consensus regularization to these domain classifiers is employed to help the network generate domain-invariant detections. We evaluate our proposed method on popular datasets like Cityscapes, KITTI, SIM10K and \etal{}. The results demonstrate significant improvement when tested under different cross-domain scenarios.
\end{abstract}

\section{Introduction}
Object detection aims to localize and classify objects of interest in a given image. In recent years, tremendous successful object detection models~\cite{ren2015faster,liu2016ssd,redmon2016you} have been proposed since the appearance of deep convolutional neural network (CNN). However, a new challenge recognized as "domain shift" starts haunting the computer vision community. Domain shift is referred to as the distribution mismatch between the source and target domain which leads to performance drop. It is caused by variations in images, including different weather conditions, camera's angle of view, image quality, \etal{}. Take autonomous driving as an example, a reliable object detection model is supposed to work stably under all circumstances. However, the training data is usually collected on sunny days with clear view while in reality a car may encounter adverse weather conditions including snow and fog where visibility is compromised. Additionally, the positions of cameras may vary in test environments and thus cause viewpoint variations.

Ideally, relabeling on the target domain is the most straightforward way to solve the domain shift problem. But such handcrafted annotations come with expensive time and economic cost. With expectations for annotation-free methods, domain adaptation strives to eliminate the domain discrepancy without supervision on the target domain.

Domain Adaptation (DA) is first widely applied to classification tasks, where distance metrics like Maximum Mean Discrepancy (MMD) measure the domain shift and supervise the model to learn domain-invariant features. Later, adversarial training strategy using domain classifiers and Gradient Reversal Layer (GRL) proved to be a more effective method to learn robust cross-domain features. During training phase, the domain classifier becomes gradually better at distinguishing source and target domain data while the backbone feature extractor learns to generate more domain-indistinguishable features. Finally, the feature extractor is able to produce domain-invariant features.

DA for object detection inherits and extends the same adversarial training idea. Similar to DA for classification , DA for detection employs adversarial training for the backbone feature extractor. However, beyond classification, object detectors need to localize and classify each object of interest. As a result, an extra domain classifier classifying each instance feature is employed to urges the feature extractor to be domain-invariant at an instance level.

\begin{figure}
    \centering
    \includegraphics[scale=0.45]{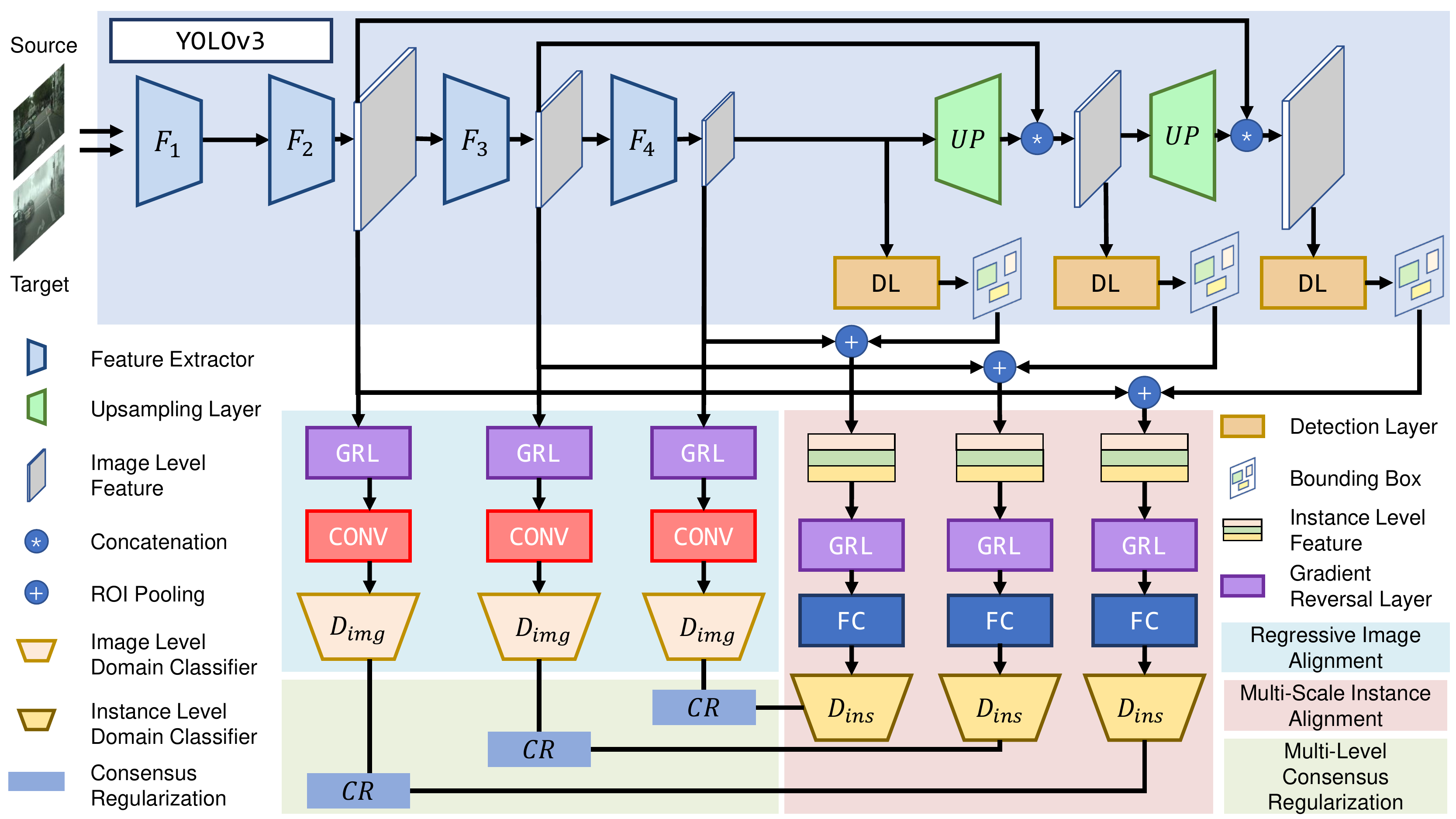}
    \caption{The architecture of our proposed DA-YOLO model. The upper part is the original YOLOv3. The lower part consists of the three domain adaptation modules RIA, MSIA and MLCR. RIA and MSIA use three domain classifiers to perform image and instance level adaptation respectively. MLCR enforce consensus between image and instance level domain classifiers to encourage the network to generate domain-invariant detections.}
    \label{fig:da_yolo}
\end{figure}

This line of adversarial detection adaptation was pioneered by~\cite{chen2018domain}, who use Faster R-CNN as their base detector model. Following researches adhered to this convention and Faster R-CNN became the de facto detector for domain adaptation. In addition, the two-stage characteristic of Faster R-CNN makes it ideal to apply domain adaptation on instance level features. It's convenient for domain classifiers to directly use the uniform instance level features produced by the Region Proposal Network (RPN) and Region of Interest (ROI) Pooling.

Despite its popularity and convenience to take advantage of the Region Proposal Network, Faster R-CNN is not an ideal choice in real world applications where time performance is critical. Compared to Faster R-CNN, YOLO~\cite{redmon2016you}, a representative one-stage detector, is a more favorable choice because of  its amazing real-time performance, simplicity and portability. YOLOv3~\cite{redmon2018yolov3}, a popular version of YOLO, is widely used in industry, including video surveillance, crowd detection and autonomous driving. Yet, research on domain adaptive one-stage detectors remains rare.

In this paper, we introduce a novel Domain Adaptive YOLO (DA-YOLO) that performs domain adaptation for one-stage detector YOLOv3. The overall architecture of this model can be viewed in Fig.\ref{fig:da_yolo}. First, we propose Regressive Image Alignment (RIA) to reduce domain discrepancy at image level. RIA uses three domain classifiers at different layers of the YOLOv3 feature extractor to predict the domain label of the feature maps. Then, it employs adversarial training strategy to align image level features. By assigning different weights to these image level domain classifiers, RIA strictly aligns local features and loosely aligns global ones. Second, we propose Multi-Scale Instance Alignment (MSIA) for instance level domain adaptation. Without region proposals that is available in two-stage detectors , MSIA exploits the YOLOv3’s three scale detections. MSIA uses three domain classifiers for these detections to align the instance level features. Finally, we apply Multi-Level Consensus Regularization (MLCR) to the domain classifiers to drive the network to produce domain-invariant detections.

To summarize, our contributions in this paper are threefold: 1) We design two novel domain adaptation modules to tackle the domain shift problem 2) We propose a paradigm of domain adaptation for one-stage detectors. To the best of our knowledge, this is the first work proposed to unify image level and instance level adaptation for one-stage detectors. 3)We conduct extensive domain adaptation experiments using the Cityscapes, Foggy Cityscapes, KITTI, SIM10K datasets. Results demonstrate effectiveness of our proposed Domain Adaptive YOLO in various cross-domain scenarios.

\section{Related Works}
{\bf Object Detection:} Object detection methods thrive with the application of deep neural network. They can be roughly divided into two categories: two-stage methods and one-stage methods. R-CNN series~\cite{girshick2014rich,girshick2015fast, ren2015faster} are representatives of two-stage detectors which first generate proposals and then classify them. Meanwhile, YOLO is the representative one-stage detector and becomes the widely-used one thanks to its real time performance. YOLOv2~\cite{redmon2017yolo9000} and YOLOv3~\cite{redmon2018yolov3} were introduced as incremental improvements where effective techniques like residual block are integrated. YOLOv4~\cite{bochkovskiy2020yolov4} is a combination of various tricks which achieves optimal speed and accuracy. YOLO framework remains the popular choice for industrial applications in which domain adaptation methods are much-needed.

{\bf Domain Adaptation:} Domain Adaptation aims to improve model performance on target domain with annotated source data. It was first applied to classification task by matching marginal and conditional distribution of the source and target domain. Pioneer works include TCA~\cite{pan2010domain}, JDA~\cite{long2013transfer}, JAN~\cite{long2017deep}. With the appearance of Generative Adversarial Network~\cite{goodfellow2014generative} (GAN), adversarial training strategy became popular because of its effectiveness. This strategy turns out to be very helpful in learning domain-invariant features and leads to a line of adversarial domain adaptation research, including DANN~\cite{ganin2016domain}, DSN~\cite{bousmalis2016domain}, SAN~\cite{cao2018partial} and so on~\cite{zhong2019source,hu2019multi,hu2020discriminative,zhong2019transfer,hu2020unsupervised}.

{\bf Domain Adaptation for Object detection:} Adversarial domain adaptation for object detection is explored by Domain Adaptive Faster R-CNN~\cite{chen2018domain}, which uses a two-stage detector Faster R-CNN. Subsequent researchs~\cite{he2019multi,saito2019strong,xie2019multi,wang2019few,shen2019scl} followed this two-stage paradigm and made considerable improvements. Despite its convenience for domain adaptation, a two-stage detector is of rare use in industrial applications. Instead, one-stage detectors are more favorable in practice because of its incomparable speed performance. Hence, it is of great importance to combine one-stage detectors with domain adaptation, but little attention has been paid to relative research. This situation motivates us to develop the proposed work in this paper.

There is limited study on domain adaptation for one-stage detector. YOLO in the Dark~\cite{sasagawa2020yolo} adapts YOLO by merging several pre-trained models. MS-DAYOLO~\cite{hnewa2021multiscale} employs multi-scale image level adaptation for YOLO model. However, it does not consider instance level adaptation which is proved to be equally or more important. Instance features adaptation is a more challenging task, because in one-stage detectors, region proposals are not as available as they are in two-stage detectors. In this paper, we resolve this problem by using the detections of YOLO.

\section{Methodology}
\subsection{Problem Definition}
The goal of domain adaptation is to transfer knowledge learned from the labeled source domain $D_{s}$ to the unlabeled target domain $D_{t}$. The distribution of $D_t$ is similar to $D_s$ but not exactly the same. The source domain is provided with full annotations and denoted as $D_{s} = \{(x_{i}^{s},y_{i}^{s},b_{i}^{s})\}_{i}^{n_s}$, where  $b_{i}^{s} \in R^{k \times 4}$ represents bounding box coordinates of image data $x_{i}^{i}$, $y_{i}^{s} \in R^{k \times 1}$ represents class labels for corresponding bounding boxes. Correspondingly, the target domain has no annotations, which is denoted as $D_{t} = \{(x_{j}^{t})\}_{j}^{n_t}$. By using labeled data $D_{s}$ and unlabeled data $D_{t}$, source detector can generalizes well to the target domain.

The joint distribution of source domain and target domain are denoted as $P_S(C,B,I)$ and $P_T(C,B,I)$ respectively, where $I$ stands for image representation, $B$ for bounding box and $C \in \{1,...,M\}$ for class label of an object ($M$ is the total number of classes). The domain shift originates from the joint distributions mismatch between domains, i.e., $P_S(C,B,I) \neq P_T(C,B,I)$. According to~\cite{chen2018domain}, the joint distribution can be decomposed in two ways: $P(C,B,I) = P(C,B|I)P(I)$ and $P(C,B,I) = P(C|B,I)P(B,I)$. By enforcing $P_{S}(I) = P_{T}(I)$ and $P_{S}(B,I) = P_{T}(B,I)$, we can alleviate the domain mismatch on both image level and instance level.

\subsection{ A Closer Look at Domain Adaptive Object Detection}
As the pioneer work of adversarial detection adaptation, Domain Adaptive Faster R-CNN~\cite{chen2018domain} proposes 1) image alignment, which concentrates on bridging the domain gap caused by image level variation, such as different image quality and illumination; 2) instance alignment, which focuses on reducing instance level domain shift caused by instance level variation, like difference in object size; 3) consistency regularization, which is designed for reinforcing domain-invariant localization ability.

Though such a paradigm is effective, Faster R-CNN based domain adaptation would not be well applied to real world applications. The reasons are twofold. First, two-stage detectors like Faster R-CNN require the training of backbone, RPN and detection head. The set up is neither convenient nor straightforward. Second, the time performance of two-stage detectors is unsatisfactory. For example, the state-of-the-art two-stage detector, Detectron2, implemented by Facebook AI Research can barely reach real time performance.

On the contray, one-stage detectors like YOLO have been extensively used in industrial application for its superiority in practice. For example, PP-YOLO~\cite{long2020pp} is widely used for pedestrian detection, car detection and product quality inspection. One-stage detectors are easy to use, free to customize and can achieve high performance in terms of time cost and computation cost. 

The representative one-stage detector YOLOv3,  consist of two parts: the backbone feature extractor Darknet-53 and three detection layers of different scales. The network architecture is illustrated in the upper part of Fig.\ref{fig:da_yolo}. The feature extractor takes images as inputs and provides three feature maps of different sizes to three detection layers respectively. As a result, detection outputs are generated at three different scales and combined together as final output. The training loss of YOLOv3 is composed of localization loss, classification loss and confidence loss:
\begin{small}
\begin{equation}
    \begin{split}
        \mathcal{L}_{det} &= \lambda_{coord} \sum_{i=0}^{S^2} \sum_{j=0}^{B} \mathbbm{1}_{ij}^{obj} [(x_i - \hat{x}_i)^2 + (y_i - \hat{y}_i)^2]
+ \lambda_{coord}\sum_{i=0}^{S^2} \sum_{j=0}^{B} \mathbbm{1}_{ij}^{obj} [(\sqrt{w_i} - \sqrt{\hat{w}_i})^2 + (\sqrt{h_i} - \sqrt{\hat{h}_i})^2]\\
&+ \sum_{i=0}^{S^2} \sum_{j=0}^{B}\mathbbm{1}_{ij}^{obj}(C_i - \hat{C}_i)^2 + \lambda_{coord}\sum_{i=0}^{S^2} \sum_{j=0}^{B}\mathbbm{1}_{ij}^{noobj}(C_i - \hat{C}_i)^2
+ \sum_{i=0}^{S^2}\mathbbm{1}_{i}^{obj} \sum_{c\in classes}{(p_{i}(c)-\hat{p}_{i}(c))}^2
    \end{split}
\end{equation}
\end{small}
The first two terms are localization loss, next two are classification loass and the last one is confidence loss. Interested readers are referred to~\cite{redmon2016you} for further details.

\subsection{Domain Adaptive YOLO}
\subsubsection{Regressive Image Alignment}
Image level alignment proves to be an effective domain adaptation method in~\cite{chen2018domain}. However, due to gradient vanishing, it fails to elimate the domain shift sufficiently by only aligning the feature from the final feature map. Consequently,~\cite{xie2019multi} and ~\cite{hnewa2021multiscale} both propose to employ extra domain classifiers on intermediate feature maps, which proves to be a valid approach. However, as pointed out in~\cite{saito2019strong}, strong matching for features with a large receptive field, the latter part features of the feature extractor, is likely to result in negative transfer when dealing with huge domain shift.

To solve this problem, we propose Regressive Image Alignment (RIA). We first apply multiple domain classifiers to intermediate feature maps and the final feature map as previous works did. Then we assign decreasing weights to these classifiers as they take deeper feature maps as input. The RIA loss can be written as:
\begin{equation}
{\mathcal{L}}_{RIA} = - \sum_{i,k,u,v} \lambda_{k}[D_{i} \text{log} \space f_{k}(\Phi_{i,k}^{u,v}) + (1 - D_{i}) \text{log} (1 - f_{k}(\Phi_{i,k}^{u,v}))]
\end{equation}
where $\Phi_{i,k}^{(u,v)}$ represents the activation located at $(u,v)$ of the corresponding $k\text{-th}$ feature map of the $i\text{-th}$ image, $f_{k}$ denotes the domain classifier, $D_i$ is the domain label of $i\text{-th}$ training image and $\lambda_{k}$ denotes the weight assigned to the domain classifier. The RIA fully adapts the backbone feature extractor while reducing possible negative transfer.

\subsubsection{Multi-Scale Instance Alignment}
~\cite{hnewa2021multiscale} is the first work to introduce adversarial domain adaptation in one-stage detectors but only primary. Because it did not take instance alignment into account which is as effective as image alignment demonstrated in~\cite{chen2018domain}. Instance alignment is a challenging task for one-stage detectors because instance features are not at disposal like they are in two-stage detectors. We propose Multi-Scale Instance Alignment (MSIA) to resolve this challenge. Specifically, We take detection results from YOLOv3's  three different scale detection layers and use them to extract instance level features from their corresponding feature maps by ROI pooling. With access to instance features, we can incorporate instance alignment loss, which can be written as:
\begin{equation}
\mathcal{L}_{MSIA} = - \sum_{i,j,k} \lambda_{k}[D_{i} \text{log} \space p_{i,j}^{k} + (1 - D_{i}) \text{log} (1 - p_{i,j}^{k})]
\end{equation}
where $p_{i,j}^{k}$ denotes the probability output of the $j\text{-th}$ detection at the $k\text{-th}$ scale in the $i\text{-th}$ image. Aligning instance features assists to eliminate variances in appearance, shape, viewpoint between the source and target domain's objects of interest. 

\subsubsection{Multi-Level Consistency Regularization}
By employing image and instance alignment, the network is able to produce domain-invariant features. However, it is not guaranteed to produce domain-invariant detections, which is also critical for object detection. Ideally, we expect to obtain a domain-invariant bounding box predictor $P(B|I)$. But in practice, the bounding box predictor $P(B|D,I)$ is biased, where $D$ represents domain label. According to equation $(5)$ in ~\cite{chen2018domain}, we have:
\begin{equation}
P(D|B,I)P(B|I) = P(B|D,I)P(D|I)
\end{equation}
To alleviate bounding box predictor bias, it is necessary to enforce $P(D|B,I) = P(D|I)$, which is a consensus between instance level and image level domain classifiers. Consequently, $P(B|D,I)$ will approximate $P(B|I)$.

Since YOLOv3 detects object on three different scales, we propose multi-level consensus regularization to adaptively detect object on different scales, which can be written as:
\begin{equation}
\mathcal{L}_{MLCR} = \sum_{i,j,k}||\frac{1}{|I_k|} \sum_{u,v}\Phi_{i,k}^{(u,v)} - p_{i,j}^{k}||_{2}
\end{equation}

where $|I_k|$ denotes the number of activations on the $k\text{-th}$ feature map. By imposing such multi-level regularization, each detection layer of YOLO is encouraged to produce domain-invariant detections.

\subsubsection{Network Overview}
The complete architecture of our network is shown in Fig.\ref{fig:da_yolo} We build our proposed domain adaptation modules on YOLOv3 and the combination of them constitute the Domain Adaptive YOLO. Note that the shown architecture is specifically designed for training stage while the detector is the only component for testing stage.

The YOLOv3 first uses a series of feature extractors to produce three feature maps of small, medium and large scale. Two sequential upsample layers take the last feature map (small scale) as input and produce a new medium and large feature map which are concatenated with previous ones. The latest three features maps are fed into detection layers and detection results are generated.

The RIA module takes the three scale feature maps as input and uses domain classifiers to predict their domain label. The MSIA module uses different scale detections to extract instance level features and feed them to domain classifiers too. The domain classification losses for RIA and MSIA are calculated in $eq.(2)$ and $eq.(3)$ to adapt the network. The corresponding image level and instance level domain classifiers are finally regularized by the MLCR module to supervise the network to generate domain-invariant detections.

The complete training loss is as follow:
\begin{equation}
\mathcal{L} = \mathcal{L}_{det} + \lambda(\mathcal{L}_{RIA} + \mathcal{L}_{MSIA} + \mathcal{L}_{MCLR})
\end{equation}
where $\lambda$ is the hyperparameter to balance the impact of the domain adaptation loss. This domain adaptation loss is reversed by the GRL to carry out the adversarial training.

\section{Experiments}
In this section, we evaluate our proposed model DA-YOLO in three domain adaptation scenarios: 1) from clear to foggy: where the source domain is photos collected in sunny weather and the target is in foggy weather. 2) from one scene to another: where the source and target domains contain photos taken by different cameras in different scenes. 3) from synthetic to real: where the source domain is images from computer games and the target domain is form real world.

\subsection{Datasets  and Protocols}
{\bf Cityscapes:} Cityscapes~\cite{cordts2016cityscapes} collects urban street scenes in good/medium weather condition across 50 different cities. It has 5,000 annotated images of 30 classes.
% Though it is not intended for object detection at the beginning, it is extended by generating bounding boxes from pixel-level segmentation masks.

\noindent
{\bf Foggy Cityscapes:} Foggy Cityscapes~\cite{sakaridis2018semantic} simulates foggy scenes using the same images form Cityscapes, making it ideal for domain adaptation experiments. As a result, it inherits the same annotations from Cityscapes.

\noindent
{\bf KITTI:} KITTI~\cite{geiger2015kitti} collects images by driving in a mid-size city, Karlsruhe, in rural areas and on highways. It is based on a autonomous driving platform. It has 14999 images and includes classes like person and car. In our experiment, we only use the 6684 training images and annotations for car.

\noindent
{\bf SIM10K:} SIM10K~\cite{johnson2016driving} collects synthetic images from a video game called Grand Theft Auto V (GTA V). It has a total of 10,000 images and annotations mostly for car.

We report Average Precision (AP) for each class and mean Average precision (mAP) with threshold of 0.5 for evaluation. To validate our proposed method, we not only report the final results of the network but also the results on different variants (RIA, MSIA, MLCR). We use the original YOLOv3 as a baseline, which is trained on source domain data without employing domain adaptation. We show the ideal performance (oracle) by training the YOLOv3 using annotated target domain data. We also compare our results with present SOTA methods based on Faster R-CNN, including~\cite{chen2018domain,he2019multi,xie2019multi,saito2019strong}.

\subsection{Experiments Details}
The experiments follow the conventional unsupervised domain adaptation setting in~\cite{chen2018domain}. The source domain is provided with full annotations while the target domain is not. Each training batch consists of one image from the source domain and one from the target. Each image is resized to width 416 and height 416 to fit the YOLOv3 network. Our code is based on the PyTorch implementations of YOLOv3 and Domain Adaptive Faster R-CNN. The network is initialized with pretrained weights before performing adaptation and all hyper-parameters remain default of the two aforementioned implementations. Specifically, learning rate of 0.001 for the backbone feature extractor and 0.01 for rest layers is used. A standard SGD algorithm of 0.0005 weight decay is applied.
\begin{table*}[tb!]
\begin{center}
\resizebox{0.7\textwidth}{!}{
\begin{tabular}{c|c|c}
\hline
car AP & KITTI $\rightarrow$ \text{Cityscapes} & SIM10K $\rightarrow$ \text{Cityscapes}\\
\hline\hline
YOLOv3 & 36.4 & 46.4 \\
\hline\hline
DAF~\cite{chen2018domain} & 38.5 & 39.0 \\
\hline
MADAF~\cite{he2019multi} & 41.1 & 41.0 \\
\hline
MLDAF~\cite{xie2019multi} & - & 42.8 \\
\hline
STRWK~\cite{saito2019strong} & - & 47.7 \\
\hline
Ours & \textbf{54.0} & \textbf{50.9} \\
\hline\hline
Oracle & 57.8 & 57.8 \\
\hline
\end{tabular}
}
\end{center}
\caption{Comparisons with other Faster R-CNN based models. We show results of domain adaptation from KITTI to Cityscapes and from SIM10K to Cityscapes. Only car AP is reported because car is the most common class.}
{\label{table1}}
\end{table*}

\begin{table*}[tb!]
\begin{center}
\resizebox{1\textwidth}{!}{
\begin{tabular}{c|cccc|cccccccc|c}
\hline
 & EIA & MSIA & MLCR & RIA & person & rider & car & truck & bus & train & mcycle & bicycle & mAP \\
\hline\hline
YOLOv3 & & & & & 25.1 & 16.7 & 44.0 & 10.1 & 29.2 & 7.6 & 10.1 & 20.8 & 32.5 \\
\hline
\multirow{4}{*}{\centering Ours} & $\checkmark$ & & & & 27.4 & 22.1 & 42.7 & 14.5 & 23.4 & 9.1 & 6.1 & 21.5 & 33.0 \\
& $\checkmark$ & $\checkmark$ & & & 28.0 & 17.5 & 44.3 & 15.6 & 23.7 & \textbf{9.1} & 7.2 & 18.7 & 33.4 \\
& $\checkmark$ & $\checkmark$ & $\checkmark$ & & 28.3 & 25.7 & 44.4 & \textbf{15.2} & 22.3 & 1.5 & \textbf{15.3} & 22.7 & 34.6 \\
& & $\checkmark$ & $\checkmark$ & $\checkmark$ & \textbf{29.5} & \textbf{27.7} & \textbf{46.1} & 9.1 & \textbf{28.2} & 4.5 & 12.7 & \textbf{24.8} & \textbf{36.1} \\
\hline\hline
Oracle & & & & & 29.9 & 19.0 & 53.4 & 17.8 & 26.1 & 10.6 & 11.3 & 18.7 & 38.4\\
\hline
\end{tabular}
}
\end{center}
\caption{Quantitative results of domain adaptation from Cityscapes to Foggy Cityscapes. The average precision (AP) of each class and the overall mean Average Precision (mAP) is reported. EIA denotes image alignment with equal weight for each image level domain classifiers. MSIA denotes the MSIA module. MLCR denotes the MLCR module. RIA denotes image alignment with decreasing weight for the three image level domain classifiers, which is the RIA module.}
{\label{table2}}
\end{table*}

\begin{table*}[tb!]
\begin{center}
\resizebox{0.5\textwidth}{!}{
\begin{tabular}{c|cccc|c}
\hline
& EIA & MSIA & MLCR & RIA & car AP \\
\hline\hline
YOLOv3 & & & & & 36.4 \\
\hline
\multirow{4}{*}{\hfil Ours} & \checkmark & & & & 42.2 \\
& \checkmark & \checkmark & & & 50.0 \\
& \checkmark & \checkmark & \checkmark & & 51.8 \\
& & \checkmark & \checkmark & \checkmark & \textbf{54.0} \\
\hline
Oracle & & & & & 57.8\\
\hline
\end{tabular}
}
\end{center}
\caption{Quantitative results of domain adaptation from KITTI to Cityscapes.}
{\label{table3}}
\end{table*}

\subsection{Results}
Table \ref{table1} compares our model with other Faster R-CNN based ones. We perform this evaluation in two domain adaptation settings, KITTI to Cityscapes and SIM10K to Cityscapes. Though DAF~\cite{chen2018domain} is effective, it only conducts image level alignment on the final feature map of Faster R-CNN. MADAF~\cite{he2019multi} and MLDAF~\cite{xie2019multi} extend DAF and align image level feature at different level of the backbone feature extractor. STRWK~\cite{saito2019strong} strongly aligns local feature and weakly align global features. They all didn't effectively extend the instance alignment module. Our proposed method conducts multi-level alignment on both image and instance levels, which achieves 17.6\% and 4.5\% performance gain and precedes other models.

\begin{figure*}[tb!]
\centering
\subfigure[Raw image]{\label{fig:DaNN_t}
\begin{minipage}[t]{0.32\linewidth}
\centering
\includegraphics[width=1.9in]{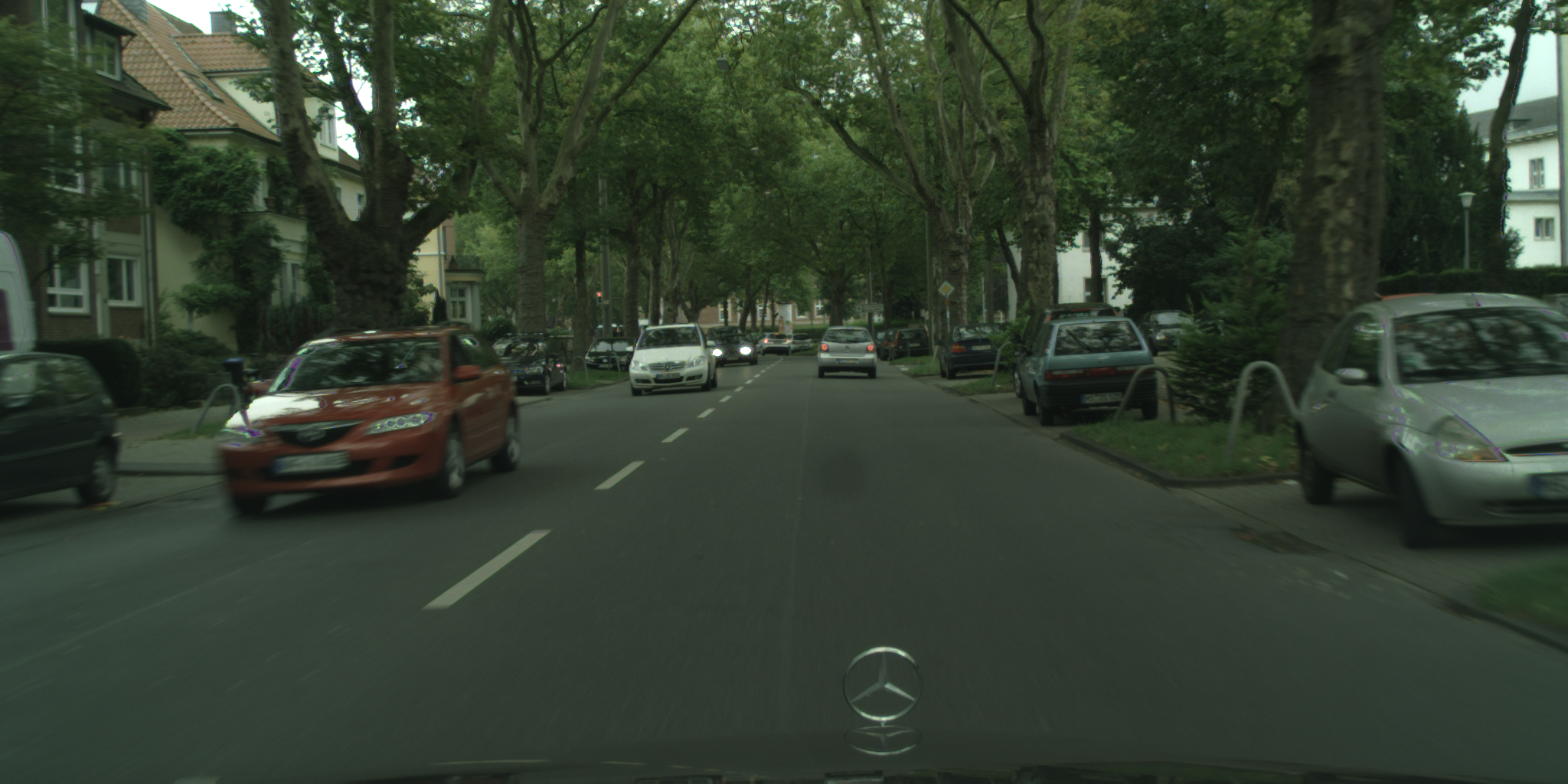}
\end{minipage}%
}%
\subfigure[Baseline (Source only)]{\label{fig:PADA_t}
\begin{minipage}[t]{0.32\linewidth}
\centering
\includegraphics[width=1.9in]{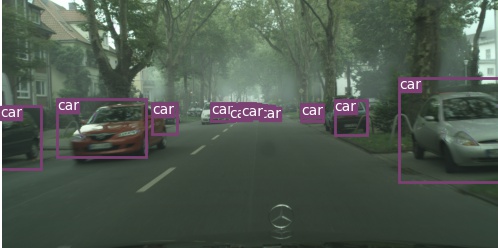}
\end{minipage}
}%
\subfigure[Proposed]{\label{fig:DPDAN_t}
\begin{minipage}[t]{0.32\linewidth}
\centering
\includegraphics[width=1.9in]{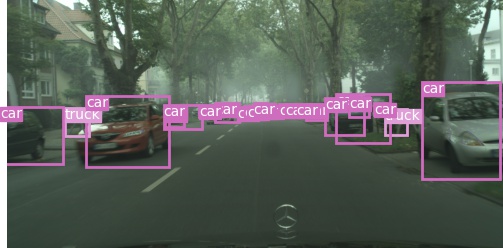}
\end{minipage}
}%
% \vspace{-10pt}
\caption{Comparison of detection result between baseline model and our proposed method on a raw image.}
\label{fig:examples}
\end{figure*}

\subsection{Analysis}
\textbf{Ablation Study:} We carry out ablation study in two domain adaptation scenarios, Cityscapes to Foggy Cityscapes and KITTI to Cityscapes, to verify the  efficacy of our three proposed modules. Performance results are summarized in Table \ref{table2} and Table \ref{table3}. In the task from Cityscapes to Foggy Cityscapes, by applying image alignment with equal weight to each image level domain classifier (EIA), we achieve 0.5\% performance gain. Aggregating MSIA and MLCR further can bring 0.4\% and 1.2\% improvement, which means these two module are effective. Finally, employing RIA instead of EIA, we have 1.5\% promotion. This validates the remarkably effectiveness of the regressive weight assignment for image level adaptation. Compared with normal multi-level image alignment, RIA assigns decreasing weight to image level domain classifiers. This will strongly match local features which is more important for domain adaptation.

In the task from KITTI to Cityscapes, similar results have been achieved. Specifically, it obtains 5.8\%, 13.6\%, 15.4\%, 17.6\% performance gain respectively for accumulating each module. The efficacy of each module is validated again.

\textbf{Detection results:} Fig.\ref{fig:examples} shows an example of detection results. In the figure, the baseline model (trained only on source domain data) misses a few cars but the proposed model can correctly detect them.

\begin{figwindow}[0,r,%
{\includegraphics[width=200pt]{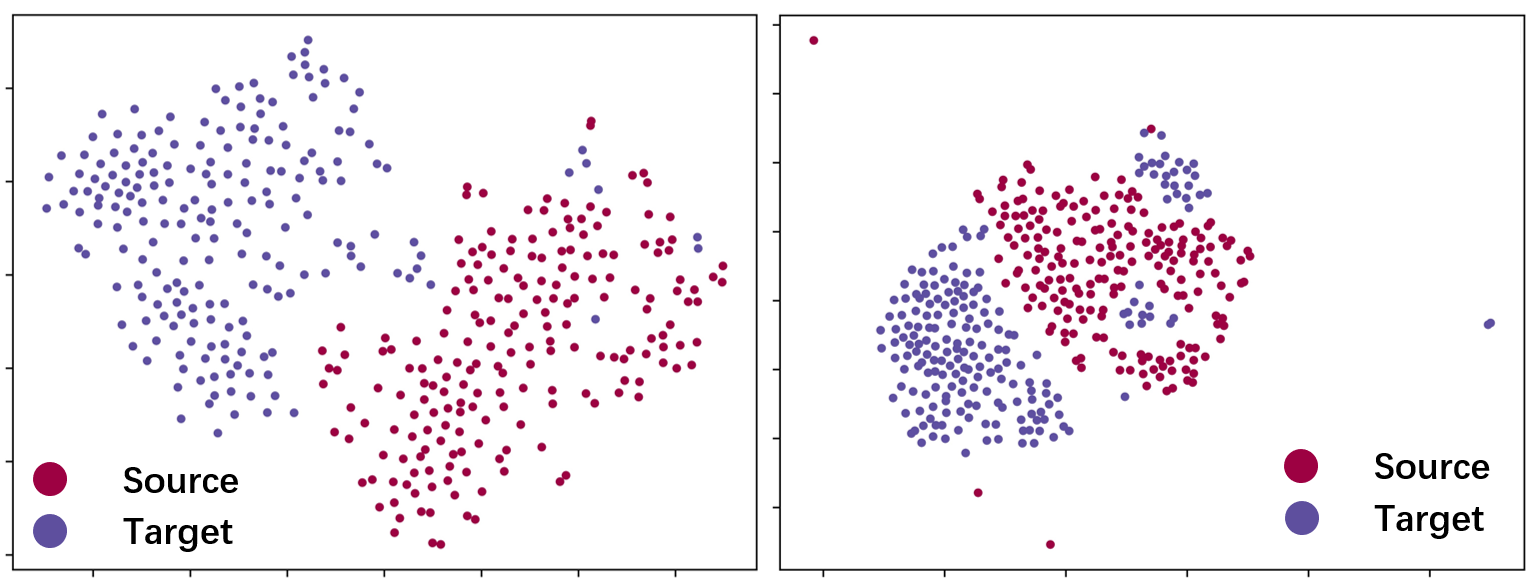}},%
{\label{fig:tsne}:Visualization of feature map (after $ F_4$) using t-SNE algorithm. (left) global feature map from source-only model (right) global feature map from DA-YOLO model.}
]
\textbf{T-SNE:} As shown in Fig.\ref{fig:tsne}, we visualize the image level features in the domain adaptation from Cityscapes to Foggy Cityscapes. Red and blue denotes source and target domain respectively. We can see that global features are aligned well compared to the source-only model (trained on source domain without domain adaptation).
\end{figwindow}

\vspace{5mm}
\section{Conclusion}
In this paper, we propose DA-YOLO for effective one-stage cross-domain adaptation. Compared with previous methods, we build our domain adaptation model on one-stage detector. Furthermore, we successfully introduce instance level adaptation for one-stage detector. Sufficient experiments on several cross-domain datasets illustrate that our method outperform previous methods that are based on Faster R-CNN and the three proposed domain adaptation modules are all effective.

\bibliography{acml21}

\end{document}